# Differentiable modeling to unify machine learning and physical models and advance Geosciences

Chaopeng Shen[1*], Alison P. Appling[2], Pierre Gentine[3], Toshiyuki Bandai[4], Hoshin Gupta[5], Alexandre Tartakovsky[6], Marco Baity-Jesi[7], Fabrizio Fenicia[7], Daniel Kifer[8], Li Li[1], Xiaofeng Liu[1], Wei Ren[9], Yi Zheng[10], Ciaran J. Harman[11], Martyn Clark[12], Matthew Farthing[13], Dapeng Feng[1], Praveen Kumar[6,14], Doaa Aboelyazeed[1], Farshid Rahmani[1], Hylke E. Beck[15], Tadd Bindas[1], Dipankar Dwivedi[16], Kuai Fang[17], Marvin Höge[7], Chris Rackauckas[18], Tirthankar Roy[19], Chonggang Xu[20], Binayak Mohanty[21], Kathryn Lawson[1]

[1] Civil and Environmental Engineering, The Pennsylvania State University, University Park, PA, USA.
[2] U.S. Geological Survey, Water Mission Area, Integrated Modeling and Prediction Division, Reston, VA, USA
[3] National Science Foundation Science and Technology Center for Learning the Earth with Artificial Intelligence and Physics (LEAP), Columbia University, New York, NY USA
[4] Life and Environmental Science Department, University of California, Merced, CA, USA
[5] Hydrology and Atmospheric Sciences, The University of Arizona, Tucson, AZ, USA.
[6] Civil and Environmental Engineering, University of Illinois, Urbana Champaign, IL, USA
[7] Eawag: Swiss Federal Institute of Aquatic Science and Technology, Dübendorf, Switzerland
[8] Computer Science and Engineering, The Pennsylvania State University, University Park, PA, USA
[9] Department of Natural Resources and the Environment, University of Connecticut, Storrs, CT, USA
[10] Southern University of Science and Technology, Shenzhen, Guangdong Province, China
[11] Department of Environmental Health and Engineering, Johns Hopkins University, Baltimore, MD, USA
[12] Global Institute for Water Security, University of Saskatchewan, Canmore, Alberta, Canada
[13] US Army Engineer Research and Development Center, Vicksburg, MS, USA
[14] Prairie Research Institute, University of Illinois, Urbana Champaign, IL, USA
[15] Physical Science and Engineering Division, King Abdullah University of Science and Technology, Thuwal, Saudi Arabia
[16] Lawrence Berkeley National Laboratory, Berkeley, CA, USA
[17] Department of Earth System Science, Stanford University, Stanford, CA, USA
[18] Computer Science and Artificial Intelligence Laboratory (CSAIL), Massachusetts Institute of Technology, Massachusetts, USA
[19] Civil and Environmental Engineering, University of Nebraska-Lincoln, NE, USA
[20] Earth and Environmental Divisions, Los Alamos National Laboratory, NM, USA
[21] Department of Biological and Agricultural Engineering, Texas A&M University, College Station, TX, USA

* Corresponding author, email cshen@engr.psu.edu

**Abstract**

Process-based modelling offers interpretability and physical consistency in many domains of geosciences but struggles to leverage large datasets efficiently. Machine-learning methods, especially deep networks, have strong predictive skills yet are unable to answer specific scientific questions. In this Perspective, we explore differentiable modelling as a pathway to dissolve the perceived barrier between process-based modelling and machine learning in the geosciences and demonstrate its potential with examples from hydrological modelling. 'Differentiable' refers to accurately and efficiently calculating gradients with respect to model variables or parameters, enabling the discovery of high-dimensional unknown relationships. Differentiable modelling involves connecting (flexible amounts of) prior physical knowledge to neural networks, pushing the boundary of physics-informed machine learning. It offers better interpretability, generalizability, and extrapolation capabilities than purely data-driven machine learning, achieving a similar level of accuracy while requiring less training data. Additionally, the performance and efficiency of differentiable models scale well with increasing data volumes. Under data-scarce scenarios, differentiable models have outperformed machine-learning models in producing short-term dynamics and decadal-scale trends owing to the imposed physical constraints. Differentiable modelling approaches are primed to enable geoscientists to ask questions, test hypotheses, and discover unrecognized physical relationships. Future work should address computational challenges, reduce uncertainty, and verify the physical significance of outputs.

**Introduction**

Geoscientific models encompass a wide range of domains, with evolving scopes and ever-increasing societal importance, especially in the face of rapid climate change. For example, hydrologic models help us manage water resources[1,2] and plan for extremes such as floods and droughts[3]; vegetation models can help predict the fate of carbon and other key biogeochemical cycles on land[4] or in the ocean[5]; agricultural models estimate crop yields and also their environmental impacts[6]; geophysical models aim to predict land surface changes via processes like landslides[7], land subsidence[8], and earthquakes; biogeochemical reactive transport models aim to understand and predict surface and subsurface water chemistry and quality[9,10]. Combining many such components, Earth System Models[11–13] and integrated assessment models[14–16] provide crucial guidance for resource managers and policy makers[17,18]. The uses of such models go beyond making predictions of the future to also facilitating communication with the stakeholders and aiding in the policy-making process[18].

Geoscientific models often share some commonalities as they describe the dynamic responses of systems to time-dependent forcings as modulated by semi-static attributes. Many such problems can be described as systems of nonlinear equations, algebraic differential equations, or ordinary and/or partial differential equations (ODE/PDEs), along with parameterizations (empirical representations) of physical processes with spatially-varying parameters. The overall system can contain multiple processes chained together, some of which are well understood while others are not[19,20]. Further, many of these process representations and parameterizations are subject to considerable uncertainty, some of which is related to scale, and thus has significant room for improvement. Here we argue that differentiable implementations of geoscientific models offer a transformative approach to simultaneously advancing process representations, parameter estimation, and predictive accuracy. In particular, differentiable implementations provide an unprecedentedly seamless connection between process-based and machine-learning-based model components, potentially enabling us to realize the value and minimize the limitations of each.

*Value and limitations of process-based geoscientific modeling*

The traditional process-based modeling (PBM) approach has served the geosciences well in helping to improve our understanding of system functions and behaviors. Due to their physical basis, they can be leveraged in hypothesis testing to assess system responses, and cause-effect relationships (see the *Physical Laws* row in Table 1), e.g., the impacts of land use changes on flooding trends[21] and future warming on glacial melt[22]. Further, they can simulate a wide variety of observed (e.g., discharge or leaf area index) and unobserved variables (e.g., groundwater recharge or fine-root carbon). Such an ability is critical to both advancing scientific understanding and to providing a narrative when communicating with the public and stakeholders, who are engaged in the decision-making process[23]. It is possible to ask and examine specific questions regarding processes within the modelled system, by progressively improving the representations of processes [24–27] and evaluating them using controlled experiments.

Despite these benefits, there remain important challenges with PBMs.

(1) Process-based models often cannot rapidly evolve with and fully exploit the information in "big data" due to the time needed to develop and test process representations and parameterizations[28,29]. Traditionally, the differences between model predictions and observations are first reconciled by parameter calibration, which adds significant uncertainty (more about this later)[30]. For model errors beyond parameter adjustments, modelers then hypothesize different causes, implement structural changes to the model, and iteratively confront the updated model hypotheses with the data[24]. This iterative process is highly expensive (in both labor and time) and dauntingly complex, and is dependent on developer intuition and legacy[31]. Consequently, it is common for the structural representation of a specific process in a geoscientific model to stagnate, with years or decades passing between structural updates[32–35].

(2) Process-based models are limited by knowledge gaps. Extensive physical, biological, and socioeconomic knowledge is required to achieve adequate representations and updates for processes in a geoscientific model, and any deficiencies can amplify errors and ambiguity. Another major challenge is the interactions of processes across disciplinary boundaries[36]. For instance, vegetation, human management, and socioeconomic systems all interact with each other and affect the water and carbon and other biogeochemical cycles[37–40]. While the intersections of these domains will continue to stimulate scientific discovery, a new paradigm could enable us to make faster progress despite knowledge gaps.

*Potential and limitations of machine-learning-based geoscientific modeling*

Irrespective of the domain of application, one cannot help but notice the "*Cambrian explosion*" of purely data-driven machine learning (ML) approaches, especially deep neural networks (NNs), applied to a wide range of scientific applications[36,41] (see Discussion A in S1, Supplementary Discussion). In geosciences, NNs have shown strong accuracy in predicting crop production[42,43], precipitation fields[44,45] and clouds[46], water quality variables[47,48] such as water temperature[49–52], dissolved oxygen[53], phosphorous[54], and nitrogen[55,56], and the full hydrologic cycle[57] including soil moisture[58–60], streamflow[61–64], evapotranspiration[65–67], groundwater levels[68], and snow[69], etc. Deep networks like long short-term memory (LSTM) networks[70], graph neural networks[63], and convolutional neural networks (CNNs)[71,72] have become widely known in geosciences. Many such studies reported noticeably better performance than conventional approaches, revealing that the latter did not fully exploit the information in the data[28] (Table S1 in Supplementary Information S1).

Nevertheless, there remain important challenges with purely data-driven ML:

(1) <u>Deep networks are data hungry.</u> The success of deep networks relies on the availability of "big data", which can, unfortunately, be sparse for many geoscientific pr oblems[56,73], where many variables are measured at dozens, hundreds, or thousands of sites only. For example, water quality data are sparse and inconsistent in temporal, spatial, and chemical coverage[74,75]. For rare and extreme events such as mega floods, droughts, and earthquakes, available data is even scarcer.

(2) <u>ML has difficulties with errors, incompleteness, or bias in the inputs or observations.</u> The quality of ML models is limited by the quantity, diversity, and quality of training data[52,76]. Since a purely data-driven model can, at best, nearly-perfectly replicate the patterns in the training data, it invariably inherits various issues from the training data including implicit or explicit biases, inadequate spatiotemporal resolutions (e.g., with satellite-based observations), and the inability to account for non-stationarity in time series due to the short data record.

(3) <u>Neural Networks remain challenging to interpret.</u> Although explainable AI methods such as layerwise relevance backpropagation[77–79] can be highly helpful in revealing some of the internal workings of a network and should be pursued, they are not designed to flexibly query a model or identify missing physics.

(4) <u>Purely data-driven ML models cannot predict untrained variables</u> (those not provided as training targets). Due to their very nature, ML-based models are designed to only output the training targets. They cannot provide an account of how events unfolded, *e.g.*, the ability to state that "*the flood occurred because the soil was saturated*" in a study where soil moisture is unobserved. This hinders both formation of hypotheses and communication with stakeholders.

(5) <u>Most geoscientific ML algorithms capture correlations and not causality</u> regarding both attributes and temporal changes. There are always confounding covarying factors in data, so that ML models can produce the "right" results for the wrong reasons, potentially making projections less reliable when circumstances are changed.

**The root of deep network's success – Differentiable Programming**

Considering both the exceptional successes and limitations of ML and especially NNs, one can ask:

*What are the foundational strengths of NNs?*

*How can we maximize these strengths while overcoming the limitations associated with data?*

*How can we extract knowledge in an interpretable form while maintaining ML-level performance?*

In answering these questions, we argue that differentiable programming (explained below) is the computing paradigm that supports the efficient training of NNs which, in turn, can deliver many philosophically and practically transformative outcomes. Traditional modeling has been dealing with optimization problems for decades (see the *Similarity* block in Table 1). However, it is argued here that only by exploiting the power of parallelized gradient-based optimization have we been able to learn from big data and train the large numbers of weights (parameters) necessary to approximate complex unknown functions.

The ability of generic NN architectures such as CNNs, LSTMs, and attention mechanisms to approximate unknown functions has achieved desirable outcomes (Figure 1 & Table 1). First, the cost of learning a few generic architectures is lower than the significant domain expertise required by traditional models, making NNs suitable for widening access to usable predictions. Second, NNs can help in the identification of previously unrecognized linkages. Third, NN training can scale up favorably with the amount of data (in terms of accuracy, generalizability, and efficiency)[76,80], in contrast with traditional modeling where the learning may saturate after some limited calibration of parameters or functions[52].

All of these features are possible only because we can now train NNs with a large number of weights, providing a large learnable function space[81,82]. The number of weights easily exceeds the optimization capabilities of conventional algorithms. The most recent computer vision model contains two billion weights[83] and LSTM models widely employed in hydrology can contain ~500,000 weights. In contrast, traditional evolutionary[84–86], or genetic[87] or particle swarm optimization methods[88] can hardly handle more than a few dozen independent parameters (Table 1).

The computing paradigm that enables efficient training of so many parameters is *Differentiable Programming*[89,90] (Figure 1), where accurate derivatives of the model outputs with respect to inputs and/or intermediate variables can be efficiently computed. Without getting into details, this paradigm is often (but not always) enabled by ML platforms, which support reverse- or forward-mode automatic differentiation (AD)[89–91] using various approaches. Models written on these platforms can, often without much effort, be *programmatically differentiable* – even where certain operations are mathematically indifferentiable (e.g., thresholding or *if* statements), the fact that they are piecewise differentiable enables gradient computations to be performed. The chain rule can be applied to efficiently accumulate the derivatives in a process called "backpropagation"[92]. Note that differentiability is normally only needed for training, not when running the model in forward mode.

Here we expand the scope and use the term *differentiable modeling* to include any method that can produce the gradients rapidly and accurately at scale. A non-AD example is that of adjoint methods, which solve accompanying equations (called adjoint equations)[93–95] for the derivatives. AD differentiates through the code in an automated manner and is independent of the problem, while adjoint methods differentiate through the mathematical model equations and thus require manual derivations of adjoint equations for each problem[96]. Many alternative gradient estimation methods, e.g., finite differences, are intractable for any reasonably-sized NNs (10,000 weights would require 10,001 forward model evaluations) and can be challenged by stiffness. Cheaply obtained gradients allow for parameter updates via various first-order gradient-descent methods[97]. Second-order methods, such as Newton Raphson, have not gained popularity for the training of NNs due to the cost of computing the Hessian matrix. The vast majority of NNs are implemented on platforms supporting differentiable programming, while most existing PBMs are not.

Historical differences in the training of geoscientists vs ML practitioners (*Education* row in Table 1) may give the impression that ML and process-based modeling are fundamentally unrelated, but the perceived divide is more of a legacy issue now that differentiable modeling is broadly accessible. In reality, both ML and parametric physical models can be expressed in nearly identical mathematical forms (*Mathematical form* row in Table 1), and the code forms also converge when both process-based and ML components are implemented within the differentiable programming paradigm.

This leads us to the conclusion of this section: ***differentiable programming is the core distinguishing feature of neural networks, and differentiable modeling can serve as the basis for unifying NN and process-based geoscientific modeling.*** As we will discuss in the following sections, this unification requires only minor modifications to our conceptual modeling and implementation strategies, but it opens new doors to scientific discovery.

*Table 1. Similarities and differences between purely data-driven neural networks and purely process-based models. [Pro] annotates the comparative strengths, also shown in green text. In the equations, W stands for weights of the neural network g; θ stands for the physical parameters of the process-based model f; x, u and A are dynamic forcings, state variables, and semi-static attributes, respectively; and L represents the loss function which quantifies the difference between simulation outputs and observations.*

|  | **Purely data-driven neural networks** | **Purely process-based models** |
|---|---|---|
|  | *Similarities* | |
| Mathematical form | $y = g^W(u, x, A)$ <br> $W = argmin(L(y, y^*))$ | $y = f^\theta(u, x, A)$ <br> $\theta = argmin(L(y, y^*))$ |
| Programmatically differentiable | Yes | Traditionally no, but could be reimplemented on differentiable platforms or supported by new libraries |
|  | *Differences* | |
| Ease of use | [Pro] Generic model architecture – Easy to develop even without domain expertise. | Specialized domain knowledge |
| Architecture | [Pro] Generic structure with a large number of weights that allow the model to approximate a wide range of functions. | Specific structural priors representing human understanding of physics, with a small number of parameters |
| Data | [Pro] Capable of efficiently learning from and obtaining scaling benefits from big data. | Typically calibrated at a few sites, or a few parameters are calibrated in a regionalization equation. Learning saturates at a small data quantity. <br> [Pro] The potential to overcome data limitations in accuracy, resolution, and availability. |
| Training/ Calibration | [Pro] Trained using gradient descent, supported by differentiable programming. | Calibrated using various small-scale algorithms. Normally code does not support differentiable programming. |
| Unknown processes | [Pro] Data can be used to make up for processes we are not certain about. This also means we can learn unrecognized connections and expand knowledge. | We must specify the processes to be used in the model, even if they are only assumptions. |
| Outputs | Output trained variables only. | [Pro] Output many intermediate variables that facilitate providing an interpretable full narrative. |
| Physical laws | May not fully respect physical laws. | [Pro] Respect physical laws. Help us to assess cause-effect relationships. |
| Interpretation | Difficult to interpret | [Pro] Elucidate physical processes, allowing us to ask specific science questions. |
| Education | Taught in computer science or data science curricula. | Taught in engineering or science curricula. |

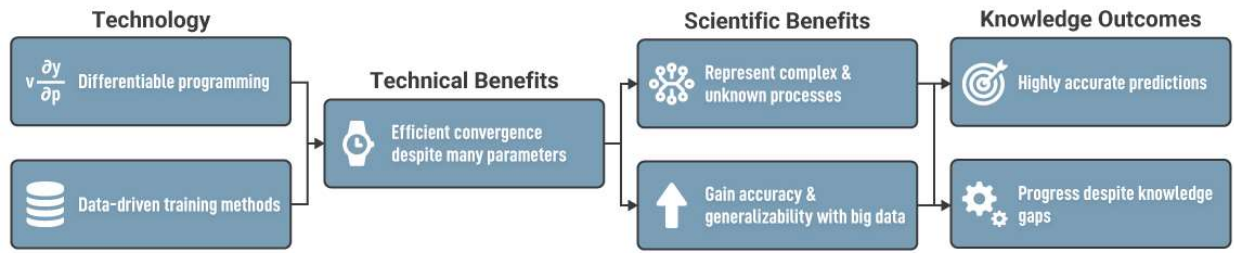

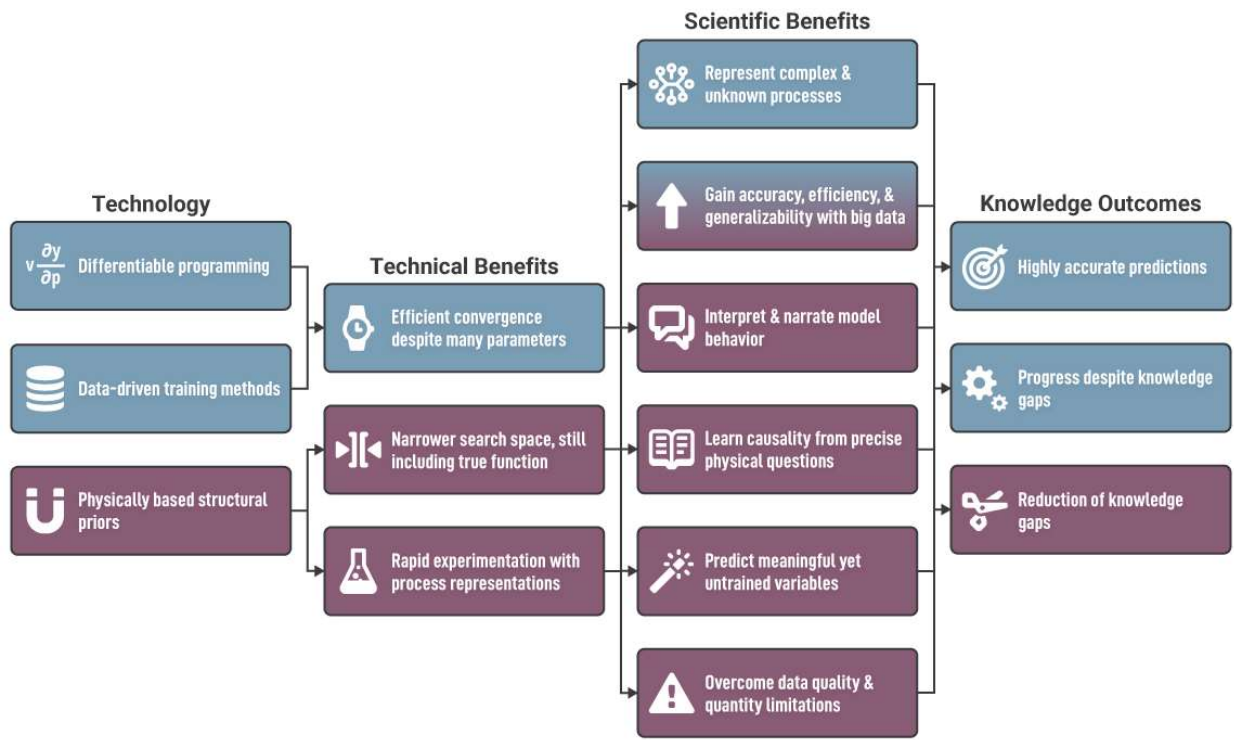

*Figure 1. (a) ML (blue boxes) gives us great results with easy-to-use models, resulting from the complexity of neural networks (many parameters) and the technologies that make it feasible to train such complex models. The most fundamental of these technologies is differentiable programming. (b) In the DG paradigm, which incorporates differentiable non-ML model components (physically based structural priors), we can now obtain additional great features (plum boxes) while retaining and augmenting the old ones (blue and blue-plum boxes, respectively).*

**Differentiable Geosciences: Absorbing the core power of scientific ML into geoscientific domains**
*What is differentiable modeling in the geosciences?*

Here we advocate for a new modeling paradigm: "*Differentiable Geoscientific modeling*", or simply "*Differentiable Geosciences*" (DG). DG refers to the use of models intermingling process-based descriptions and NNs to simulate geoscientific processes, update our physical process representations, learn physically meaningful parameters, quantify uncertainty, etc. DG allows us to replace poorly-understood or low-accuracy process-based model components with ML components that may be more accurate, while retaining those process-based model components that we already trust or want to improve. DG may also exploit gradients for other purposes such as sensitivity analysis or trajectory optimization. A distinct feature of DG is its full programmatical differentiability – that is, the whole model needs to support gradient calculation from the start to the end of the workflow – to ensure that we can incorporate neural network units that can adapt to and evolve from data. The process-based descriptions retained in the model can be called the structural priors. DG seeks to marry the core of NN models – their optimizing and learning capabilities – to geoscientific process descriptions.

DG can be considered a branch of scientific machine learning[98,99] that emphasizes improving process representations and understanding. With DG, we trade the model genericity for physical interpretability, with minimal compromises to accuracy. DG reduces the cost (in terms of data) of finding good solutions because the structural priors serve to constrain the model. Meanwhile it also scales well with data quantity and can reap the benefits of big data, just as does purely data-driven ML. There are two perspectives from which we can view DG models (Figure 2):

(a) they are ML models constrained to a smaller searchable space by the structural priors.

(b) they are PBMs augmented with learnable and adaptable components (and thus an expanded searchable space) provided by NNs.

In DG, NNs can be commissioned in a wide variety of ways, ranging from learning parameters[100] to updating assumptions used in the model (e.g., process representations)[76], and from estimating time-dependent forcing terms to describing the whole space-time solution[101]. The next section provides some forms of use cases, and examples are provided in *Classes of DG methods with examples* section below. DG is different from previous concepts of physics-guided machine learning (PGML) or not-fully-differentiable models in the methodology (must be fully differentiable), mission (to advance process understanding), and philosophy (whether treating physical law as truth or not). Please see Supplementary Information S1, Discussion C.

***From technical breakthrough to philosophical change – why will DG be transformative?***

While efficient gradient calculation may appear to be merely a technical change, it is likely to transform our modeling philosophy and scientific objectives. First, the ability to approximate complex, unknown functions greatly broadens the type of questions we can ask, by enabling us to treat trusted components as priors and focus on improving uncertain model components, one at a time. To explain this idea in concise mathematical terms, let us consider a physics-based model *y=g(u, x, θ)* where *u, x, θ* represent state variables, dynamic forcings, and physical parameters, respectively (This representation encompasses differential equations, i.e., ∂u/∂t= *g(u, x, θ)*, but is more generic). Traditional inversion algorithms only estimate the parameters, i.e., asking "*θ =?*") while requiring that the functional form *g* be assumed *a priori* (except for some rigid methods, e.g., nonparametric regression, which require complicated derivations and specialized training algorithms, and thus have not gained popularity) []. However, differentiable models allow us to ask questions about the functional form, i.e., "*g =?*", by training a neural network (*NN*) (or parameterized functions) to replace *g*: y = $NN^W(u, x, \theta)$ where *W* is the high-dimensional weights (see examples later). Hence, with DG, we now can place our question mark precisely in the model. The functions to estimate could be

(i) a parameterization scheme, as done in differentiable parameter learning[100]: $y = g(u, x, \theta = NN^W(A))$;
(ii) a module in a model, e.g., where we can replace $g_3$ in $y = g(g_1, g_2, g_3(u, x, \theta))$ with *NN*: $y = g(g_1, g_2, NN^W(u, x, \theta))$, as Feng et al.[102] optionally replaced the runoff function; or

(iii) a part of a governing equation or constitutive laws, *e.g.*, we can estimate $NN^W$ in $\partial u/\partial t = g(g_1, g_2, NN^W(u, x, \theta))$[103,104].

In the above equations, physical process equations provide a backbone for the overall model; e.g., in (i) the physical backbone is $g$; in (ii) and (iii) the physical backbone is $g$, $g_1$, $g_2$ and $g_3$. The unchanged parts (structural priors), i.e., $g$, $g_1$, $g_2$ in (ii) and (iii), critically serve as physical constraints, allowing us to isolate and focus our attention (and data) on the most unknown model components. We may gain insights by simply visualizing the relationships learned by $NN^W$ [63,105] or applying knowledge distillation methods[106]. We are also able to evolve better process representations for some model components like $g_1$ or $g_2$ mentioned above, e.g., the relation between soil moisture and effective rainfall in conceptual hydrologic models, without needing a full understanding of all the processes. This precision of questioning is opposed to some popular off-the-shelf interpretive AI approaches, e.g., layerwise relevance propagation[77,107], Shapley additive explanations[108], or local surrogate methods[109], that are limited to only asking a few fixed questions, e.g., *which parts of the inputs caused this result?* Moreover, in geoscientific modeling, directly interpreting the trained sensitivities may be risky – with only limited measurement sites, the trained relationship related to the spatial attributes tends to be overfitted.

DG provides a framework for combining deductive reasoning and inductive learning. Purely data-driven models are inductive and seek to derive almost all relationships from data, whereas process-based models first posit hypotheses and then test those hypothesis using data, albeit facing many challenges in doing so. The DG paradigm posits a user-defined number of structural priors, and then identifies many other parts of the model from data. This design follows the traditional scientific approach that identifies parsimonious models to reflect the general properties of the phenomenon, along with a quantification of the predictable aspects that are not yet understood[110]. Moreover, differentiable, learnable models can and have obtained state-of-the-art performance that can match fully data-driven models (Supplementary Information S1, Discussion B).

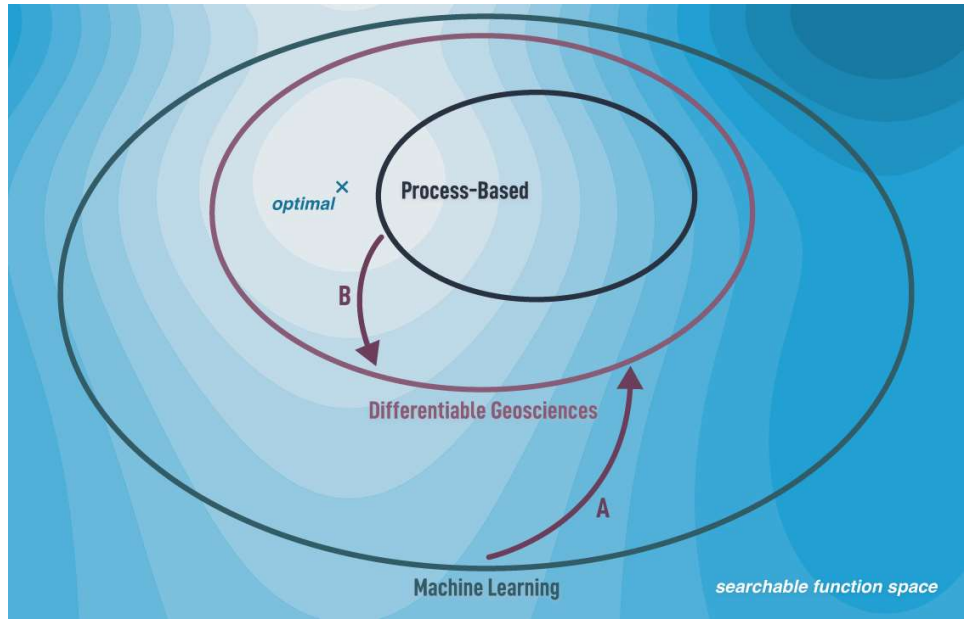

*Figure 2. Differentiable models can be viewed as (A) machine learning models guided into a smaller searchable space by structural priors or (B) process-based models with expanded search space supported by learnable units. The background fill colors indicate model optimality, related to the cost function if we had infinite data.*

### *Why is differentiable modeling particularly valuable for the geosciences?*

First of all, geoscientific data are strongly imbalanced in spatial extent, temporal coverage, and in terms of variables of interest. While satellites can measure leaf area index[111] or coarse-resolution surface soil moisture[112,113] all over the world, there are a limited number of sites measuring photosynthesis rates[114] or streamflow, especially in Africa and Asia[115]; and there is very limited knowledge of subsurface properties. Purely data-driven ML may be biased or stymied by these data limitations, which may be overcome by the inclusion of physics. Indeed, preliminary analysis shows that differentiable models with a physical model as the backbone can outperform LSTM in regional extrapolation[116].

The second major motivation is system nonstationarity induced by climate change, which could drive many systems out of the previously observed range of variability[117]. Data-driven methods are tailored to the training data and may not maintain accuracy in the face of strongly changing conditions (this is a nuanced statement as models like LSTM may have highly competitive scores even in long-term projection tests[62,116], but nonetheless experience large declines in accuracy when faced with nonstationary processes). Careful testing suggests adding stronger priors may lead to better future projections[116].

As DG models can also output any diagnostic (latent) variable available from the process-based equations within the DG model, we can perform model conditioning and/or data assimilation operations with sparse and scattered data. By conditioning, we mean constraining the model using an observable to learn more realistic parameters or processes so that the overall model dynamics are better. For example, satellite-based soil moisture data can condition a hydrologic model to better predict vegetation water use[100] or primary productivity; streamflow can constrain a model to better simulate snow water equivalent[118]. For data assimilation, the model can use recent observations of B to improve the short-term forecast of A, as B can also help to update our model state variables.

Physical parameters play key roles in geoscientific models in modulating the behaviors of the system. Parameter estimation transfers information from either (i) raw observable physiographical variables or (ii) fine-scale dynamics to parameters. Quite often, we have no ground truth information for the parameters and they require inversion using observations or high-resolution simulations. Parameter estimation has, for decades, been fraught with uncertainty, ambiguity, and frustration. Due to different parameters producing very similar output and their sensitivity to spatiotemporal resolutions, calibration at a geographic location can often lead to nonunique inference (sometimes referred to as "equifinality")[119–121]. Extending parameters to unmonitored locations requires "regionalization", which also introduces uncertainty. Because of increasing geospatial data availability, parameter estimation is an area where machine learning is well-poised to make significant progress. A novel aspect is that, as with purely data-driven ML, DG methods provide favorable scaling relationships – more training data leads to improved performance, efficiency, and generalizability[100] (discussed in Supplementary Information Text S1).

### *What are the promises of differentiable modeling in geosciences?*

We hope to evolve differentiable models so that we can gain process knowledge while improving the model predictions. Success can be claimed if we obtain models with the following features:

(i) Predictive accuracy and transferability equal to or superseding purely data-driven models for extensively measured variables;
(ii) Models capable of structural evolution, i.e., we can improve the parameterization and formulation of the processes;
(iii) Accurate generalizability to data-sparse regions or into long-term future;
(iv) Conservation of mass/energy/momentum;

(v) Consistency of internal physical fluxes and states that can provide a full narrative of the events and full support to downstream processes;
(vi) Permits efficient isolation of one uncertain model component at a time to learn physics with less ambiguity.

This wish list is ambitious and yet partial. However, as shown below, some examples already demonstrated the plausibility of these goals.

**Motivating questions for DG**

With differentiable geosciences models, we hope to ask and answer the following types of questions:
a. What is the relationship between variable x and variable y?
b. What is the missing physics as part of the differential equation?
c. What should have been the assumption or function here?
d. How does factor A influence parameter $\beta$?
e. Which process is causing phenomenon P?
f. What will happen in new environmental conditions?
g. What is the information content of datasets, either input or target data for training?

Most domains in geosciences could benefit from DG (Figure 3). To provide more concrete motivating examples, we now list one example question that DG is primed to answer from each of the domains below (ordered alphabetically):

Agriculture: *Can we predict crop phenology dynamics (e.g., planting, shooting, flowering, harvesting) and assess potential production risk under future climate change (type f), which involves interconnected biotic, abiotic, and human influences?* DG can optimize model representations of more and less understood components of this interconnected system for accuracy even in climate extrapolations.

Climate: *Can we predict cloud processes and ocean eddies and their impact on climate sensitivity?* PBM-NN hybrids implemented with DG can help to improve cloud representations.

Ecosystem: *Should we parameterize ecosystem models regarding carbon and nutrient cycles on the plant functional type level or the trait level (type c)?* Testing the configurations of differentiable parameter learning schemes could answer this question.

Coastal: *Can we better leverage emerging sensing platforms while improving our model representations of sediment transport and nonlinear wave-wave interactions in order to infer nearshore bathymetry at large scales (type g)?*

Cryosphere: *Can we leverage both physics and data to create more accurate models for ice dynamics within the cryosphere and better constrain its fate under climate change(type f)?* For example, the plumbing system for melted water and its influence on ice-basal bed rock friction are two of the key components for ice mass movement[121,122], with increasingly available data.

Coastal: *Can we better leverage emerging sensing platforms while improving our model representations of sediment transport and nonlinear wave-wave interactions in order to infer nearshore bathymetry at large scales (type g)?*

Geohazards: *Can we use space-based observations of geohazards, e.g., landslides[122], to quantify subsurface properties (type d) so we can better predict future events (type f)?* Space-based observations and combined with differentiable parameter learning provides an opportunity to inversely estimate properties like soil cohesion and friction angle which are challenging and expensive to measure.

Hydraulics: *How do we estimate floodplain hydraulic parameter values efficiently at large scales using new sensing data (type a, d)?* Estimation and inversion are most difficult problems facing the hydraulics research community, e.g., Manning's *n* for flow resistance and sediment transport rate. Another example is bathymetry which is required to run any hydraulics model but hard to observe.

Hydrology: *How does global groundwater-dominated baseflow respond to climate change (type a)? What is a proper, scale-appropriate way to parameterize groundwater storage and flow at the global scale (type c)?* For this question, we cannot answer it using a purely data-driven method, but could leverage differentiable models for the diagnosis.

Soil science: *Can we find functional forms to express soil hydraulic properties (water retention and hydraulic conductivity function) that describes non-equilibrium flow (type c or b)?*

Water quality: How and to what extent do denitrification rates vary across gradients of climate, vegetation, land use, and geology conditions (type d) and thus how do they change under different climates. Nitrate is one of the most widespread and persistent contaminants. Denitrification removes nitrate from water but the rates and extent of denitrification however depend on an array of entangled environmental factors.

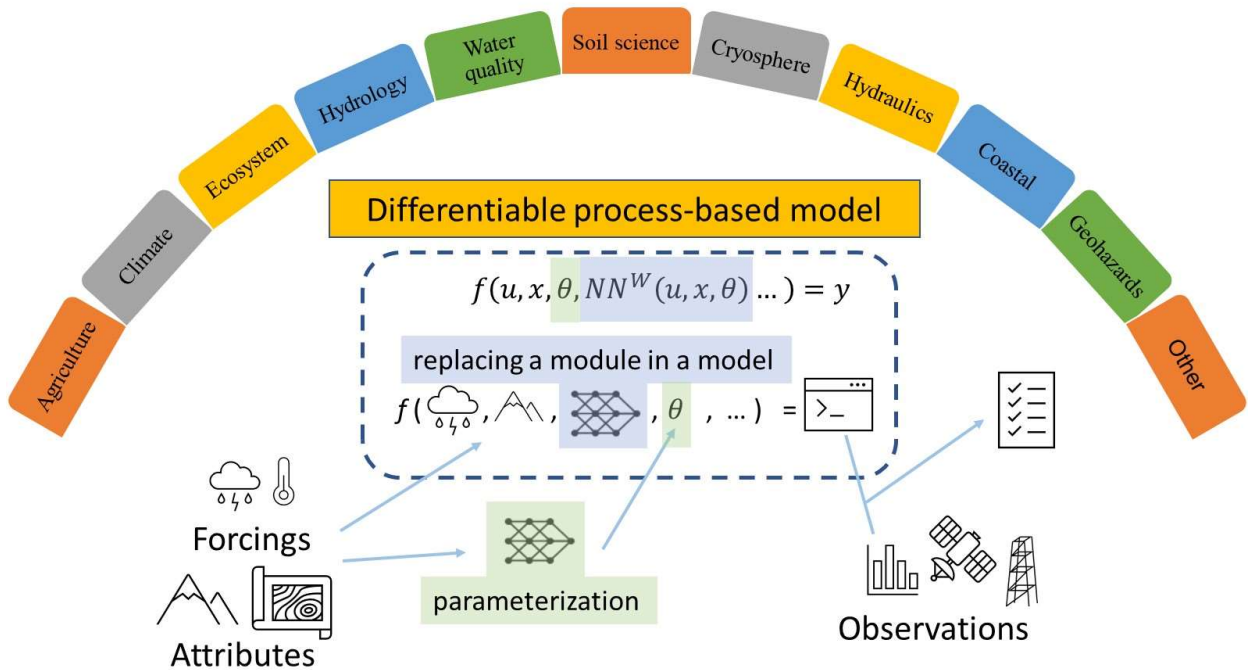

*Figure 3. Differentiable Geosciences can help almost all geoscientific domains in knowledge discovery and improving simulation quality. Green and blue highlighting is used to show how there can be multiple uses for neural networks within a single model.*

**Classes of DG methods with examples.**

DG is a young modeling paradigm that could benefit from wider participation. This section briefly describes early explorations of DG, categorized by how gradients are computed and employed. This section also gives examples, which are by no means exhaustive, to explain the concepts and to inspire more innovation.

I. Directly differentiating through numerical models and connecting them to NNs.

Among the several options, directly differentiating numerical models is the most straightforward method and is most similar to traditional models. Utilizing the AD functionality provided by modern ML platforms, one can reimplement an existing model to obtain a differentiable model version (and ensure reproducibility). Then the differentiable model is connected to NNs as discussed in the section "*Why will DG be transformative*". Because the model being trained is the same one for the forward simulation, the physics is clearly enforced, and the user can apply the forward simulator for any initial, boundary and forcing conditions. They can also migrate the learned relationships to existing implementations, e.g., the national water model, to immediately support operations. However, reimplementing a model does incur non-trivial initial development cost. Mathematical changes may be required to adapt previously non-differentiable mathematical operations to be mathematically differentiable, e.g., by replacing indexing with convolutions, and to improve parallel efficiency. While DG models may not always have to run on Graphical Process Units (GPUs), enabling GPUs will improve the computational efficiency by orders of magnitude, notwithstanding some current challenges (described in the *Challenges to address for DG* section). Our position is that in most cases, the cost is well justified due to the potential to interrogate into the model, make changes, and learn physics. The reimplementation may provide a "reset" opportunity to reexamine many of the habitually-made assumptions.

As an example, Feng et al.[102] implemented the conceptual hydrologic model HBV (a system of ODEs) on PyTorch and used coupled NNs for parameterization and optionally replaced processes with NNs (Figure 4a). Strikingly, they approached the performance level of LSTM, giving a median Nash Sutcliffe model Efficiency coefficient (NSE) of 0.732 for the CAMELS streamflow benchmark, compared to LSTM's 0.748 for the same dataset, or 0.715 vs. 0.722 for another forcing dataset (Figure 4b). They also output untrained variables such as evapotranspiration and baseflow, which agreed well with alternative estimates (Figure 4e). Moreover, in spatial extrapolation test cases, the differentiable model outperformed LSTM with respect to daily metrics and decadal trends[116] (Figure 4 c-d) due to the structural constraints, demonstrating its potential for global hydrologic modeling. Similarly, Jiang et al.[118] encoded the hydrologic model EXP-HYDRO as a recurrent NN architecture and coupled it with fully connected NNs which served as the parameterization pipeline as well as postprocessor to improve runoff. They showed that a symbiotic integration between NN and physics led to robust transferability and that snow water equivalent was well captured. In the Biogeosciences or ecosystem modeling, differentiable models found improved parameters for photosynthesis[123] at large scales.

Apart from models similar to ODEs, direct differentiation can also be applied to models operating on graphs representing the natural systems such as river networks. Bindas et al.[124] created a differentiable river routing model that was trained on daily discharge at a gauge downstream of a river network (with pretrained LSTM producing runoff as inputs to the graph) to learn a parameterization scheme for Manning's roughness coefficient ($n$). They obtained a power-law-like curve between $n$ and catchment area that was consistent with the expected $n$ behavior. Similarly, Bao et al.[125] implemented an advective dispersion equation on the river graph to simulate stream water temperature and found that the model performed better in data-sparse situations.

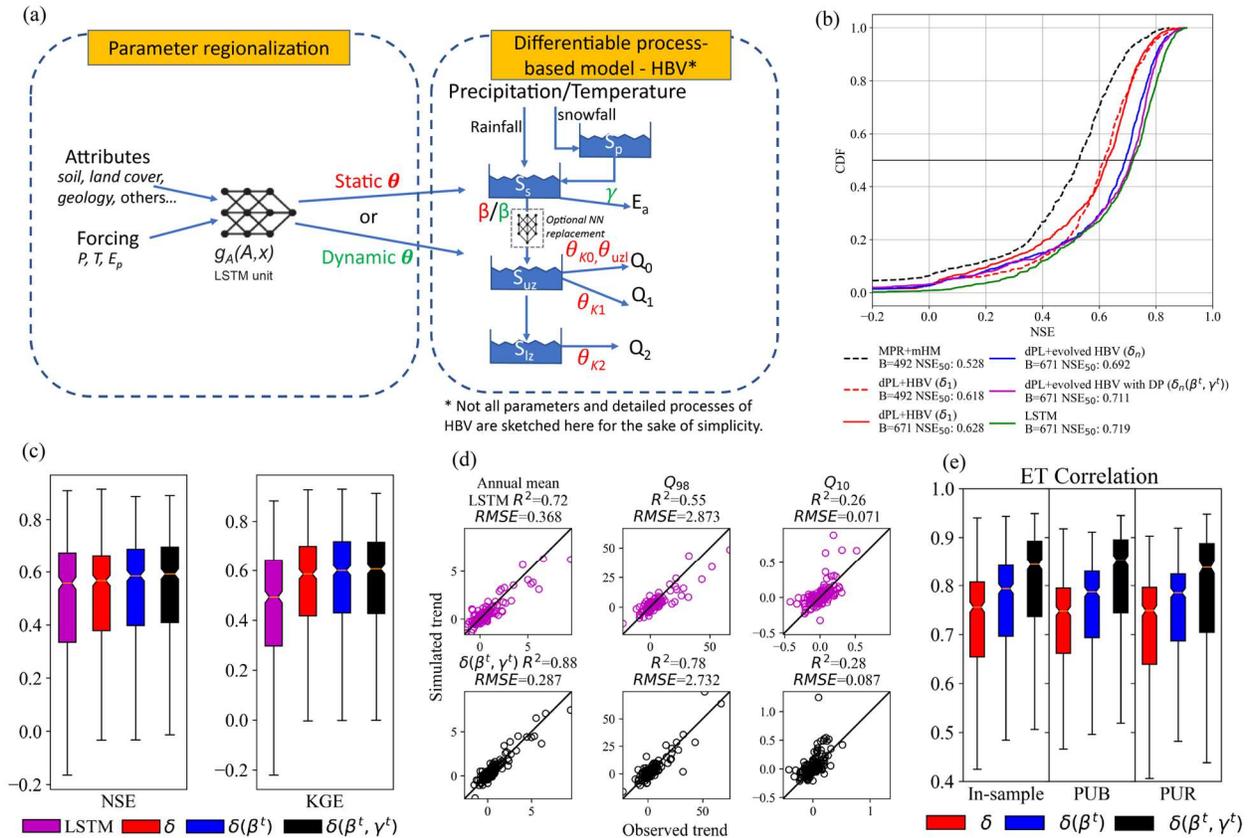

*Figure 4. (From Feng et al.[102,116]. Reprint permission obtained). (a) Sketch of a differentiable hydrologic model using process-based hydrologic model HBV as a backbone (b) For temporal test using NLDAS forcings, δ models can approach the performance of LSTM and greatly outperform traditional approaches; (c) For prediction in ungauged regions (PUR; train in some regions and test in another large ungauged region), δ models can surpass the performance of LSTM; (d) For the PUR test, δ models can predict long-term trends of annual flow percentiles more reliably than LSTM. (d) δ models can predict high-quality evapotranspiration estimate (not used in training) compared to a satellite product for both in-sample and spatial generalization tests.*

<span style="color:red">ADDITIONAL CLASSES, CHALLENGES TO DG, AND CONCLUDING REMARKS REDACTED BEFORE PAPER ACCEPTANCE.</span>

**Acknowledgements**

We attribute many ideas of the paper to a discussion in the HydroML symposium, University Park, PA, May 2022, https://bit.ly/3g3DQNX, sponsored by National Science Foundation EAR #2015680 and Penn State Institute for Computational and Data Sciences. Content related to this paper was also presented in some presentations, including Artificial Intelligence for Earth System Processes (AI4ESP) talk online https://bit.ly/3etm5aI in Nov 2021. We thank Jordan Read and James McCreight for valuable internal reviews of the manuscript. Shen was supported by National Science Foundation EAR-2221880 and Office of Science, US Department of Energy under award DE-SC0016605. Gentine acknowledges funding from the National Science Foundation Science and Technology Center, Learning the Earth with Artificial intelligence and Physics (LEAP), award #2019625 and Understanding and Modeling the Earth System with Machine Learning (USMILE) European Research Council grant. Marty Wernimont at the U.S. Geological

Survey (USGS) greatly improved the presentation of Figures 1 and 2; Wernimont and Appling were supported by the USGS Water Mission Area, Water Availability and Use Science Program. Any use of trade, firm, or product names is for descriptive purposes only and does not imply endorsement by the U.S. Government.

**Competing Interests**

KL and CS have financial interests in HydroSapient, Inc., a company which could potentially benefit from the results of this research. This interest has been reviewed by the University in accordance with its Individual Conflict of Interest policy, for the purpose of maintaining the objectivity and the integrity of research at The Pennsylvania State University.16

**Main Bibliography**

**Supplementary Information**

*S1. Supplementary Discussion*

    *A.   Recent progress in geoscientific domains with purely data-driven machine learning.*

Machine learning (ML) has gradually but pervasively permeated the vast majority of scientific disciplines, and it is transforming those sciences at an unprecedented pace. In hydrology, deep networks such as long short-term memory (LSTM) networks[70], and convolutional neural networks (CNNs)[71,72] have shown strong ability with regard to prediction of soil moisture[58–60], water supply[155], streamflow[61–64], evapotranspiration[65–67], groundwater levels[68], snow[69], and other aspects of the water cycle[57]. In water quality studies, LSTMs and CNNs have shown promise in simulating water temperature[49–52], dissolved oxygen[53], phosphorous[54], and nitrogen[55,56], among others[47,48]. In agriculture, ML approaches have been widely applied for crop production prediction[156–158]. In regional climate studies, CNN-based schemes or generative algorithms have been found to improve the forecasting of precipitation fields[44,45] and prediction of clouds (deep clouds)[46]. Often the studies have reported state-of-the-art performance when compared with conventional approaches. Typically, such high-quality predictions can be made even when a good understanding of the underlying processes is not available. We made an effort to collect a list of somewhat comparable studies with metrics for both traditional and ML models (Figure S3 and Table S1). Previous models have been highly useful in advancing science, but these results imply that they were not fully exploiting the information available in the data[28], and they can benefit from leveraging the strength of ML.

*Table S1.* ML vs. traditional model performances for a number of scientific applications with data from many sites. The metrics were computed based on simulations and observations. The lower the values, the better for RMSE, while higher is better for Pearson's correlation (COR), $R^2$, and Nash-Sutcliffe model efficiency coefficient (NSE). This is presented with many caveats, such as the ML model is optimized to match observations while traditional models have many other constraints; a selection bias – where ML did not outperform did not get published (nevertheless, one could also argue studies where PBM outperformed were not easily found). The point of this table was not to show that ML was always better, but to support the argument that ML tends to have advantages in accuracy. Also note the limitations of ML we discussed in the Introduction.

| Variable | Metric | Deep networks | Traditional | Reference |
|---|---|---|---|---|
| Stream Temperature | RMSE (°C) | 1.91 | 4.01 | Chen et al.[159] |
| | RMSE (°C) | 0.89 | 1.80 | Rahmani et al.[160] and Daraio et al.[161] |
| | Pearson COR | 0.99 | 0.91 | Rahmani et al.[160] and van Vliet et al.[162] |
| | $R^2$ | 0.942 | 0.93 | Rahmani et al.[160] |
| | NSE | 0.98 | 0.93 | Rahmani et al.[160] |
| Evapotranspiration | $R^2$ | 0.67 | 0.21 | He et al.[163] |
| | RMSE (mm/day) | 1.21 | 2.56 | |
| | NSE | 0.65 | 0.57 | Talib et al.[164] |
| Soil Moisture | RMSE | 0.027 | 0.085 | Fang et al.[58] |
| | Pearson COR | 0.87 | 0.72 | |
| | RMSE | 0.027 | 0.035 | |
| | Pearson COR | 0.87 | 0.82 | |
| | Pearson COR | 0.91 | 0.77 | Liu et al.[143] |
| | RMSE | 0.034 | 0.08 | |
| Streamflow | NSE | 0.76 | 0.68 | Seibert et al.[165] and Kratzert et al.[62] |
| | NSE | 0.9 /0.68 | - | Mohamoud and Parmar[166] |
| | Mean $R^2$ | 0.71 | - | Merritt et al.[167] |
| Dissolved oxygen | NSE | 0.78 | - | Zhi et al.[53] |
| | Median $R^2$ | - | 0.64 | Stefan and Fang[168] |
| | CC (correlation Coefficient) | 0.972 | - | Heddam[169] |
| | Median NSE | 0.760 | - | Keshtegar and Heddam[170] |

### B. *Why can differentiable process-based models achieve state-of-the-art predictive performance?*

Purely data-driven ML architectures have set a high bar for accuracy in multiple geoscience domains, such that one would be tempted to predict a loss in accuracy when adding in less-flexible process-based components. However, here it is argued that generic ML architectures are not necessary to achieve good model accuracy. As long as some model components are adaptable and learnable, we can learn from data. If we view the model as a more strongly constrained ML model (perspective "a" in Figure 2), it is easy to see that there is a potential to achieve ML-level performance if we enlarge the searchable space of PBM to include a good approximation of the true function, directed by gradient-based training. The paths we take to upgrade the models will be expert-dependent (prior-dependent), so one should not expect a unified approach at present.

Many dynamical systems in Geosciences can be written as ordinary differential equations (ODEs), e.g., rainfall runoff in a basin, crop growth, or nutrient release. While solving these equations, we run the numerical model for many steps. This is mathematically similar to recurrent neural networks, and the time integration operation is similar to the functionality achieved by some neural networks like the Residual Networks[171,172]. It should not be surprising that learnable process-based models with some ML components can perform as well as deep networks.

As we discuss in Section S1, multiple studies have already shown that differentiable, learnable models can approach the performance of purely data-driven models, or exhibit advantages in some cases where extrapolation is key. Differentiable model formulations can maintain at least two of the three desirable features: approximating complex, previously unknown functions, and the ability to assimilate information from big data. Compared to purely data-driven ML, DG trades genericity for interpretability and the ability to ask specific questions. Deep networks like CNNs, LSTMs, and transformers will be an ingrained part of DG. Eventually, deep learning will become part of the repertoire of geoscientists, just like with numerical methods[173].

### C. *How is DG related to physics-guided machine learning (PGML) and how are they different?*

Many ML-physics integration strategies with a wide variety of complexity have been proposed in the past in a seemingly scattered manner, such that a clear classification is difficult[154]. It has not been sufficiently recognized that some of these algorithms work fundamentally because they leveraged the differentiable programming tools. The scattered nature of those publications makes the landscape of ML-physics integration daunting and confusing, while hindering us from making innovations based on first principles. However, once we treat differentiability as the central tool, it serves as a compass to guide us in understanding newly proposed methods, i.e., we can ask if a method is fully (end-to-end) differentiable, how it uses gradients, how much prior information is inserted, what questions are asked, and how it scales with data. Here we outline some similarities and differences between DG and some existing methods.

DG and physics-guided (or physics-informed, theory-guided, or knowledge-guided) machine learning (PGML)[174–176] both seek to combine physics with ML, but they differ in their approaches, purposes, and philosophies. Many PGML studies seek to introduce physical constraints, e.g., as regularization or pre-training, to ML methods to gain better generalizability with less training data. PGML does not in theory need differentiable programming and partial physics could be enforced. In contrast, DG is more thorough in that it uses the numerical physical model as the backbone and demands that the entire workflow be differentiable. In terms of purposes, PGML is tasked to make the ML model more robust, while differentiable modeling seeks to update our assumptions or discover new knowledge. Relatedly, in terms of philosophies, when a physical law was introduced in PGML, it was treated as truth (albeit sometimes with some tolerance level, as in Read et al.[52]). Often, this includes all the calculations and assumptions to support the law. In DG, we do not presume the physical laws to be correct, and, rather, are constantly looking for opportunities to update existing knowledge.

There are many not-fully-differentiable methods that could be valuable for various applications but are outside of the scope of DG for this paper. For one, it is possible to incorporate ML algorithms trained offline on datasets as part of a physical model, e.g., training a neural network on turbulent heat fluxes and inserting into a hydrologic model[177]; training pedotransfer functions to infer soil parameters from soil hydraulic data[178]; training an atmospheric parameterization network on short-term cloud-resolving simulations[179]; or training ocean-mixing parameterizations on data and physical constraints[180]. While this approach has the advantage that the physical meaning of the NN is clear and stands alone, direct training data are needed for the variable of interest (thus having issues with pure ML as discussed earlier) and the network can no longer evolve and adapt in an interactive fashion, for instance to further update the model when exposed to observations. In the future these NNs could be further incorporated into DG models. Some other offline coupling methods include providing outputs of process-based models as inputs to neural networks (this helps to integrate over spatiotemporal heterogeneity)[181,182], or training ML models to predict the PBM residuals[150,183,184]. Readers are referred to Reichstein et al.[185] which promoted a number of ways to connect physics and ML for geosciences, with a brief mention of differentiable programming.

*S2. Details for some examples.*

Example 1. Part of the effort in Tsai et al.[100], which proposed differentiable parameter learning (dPL), connected the Variable Infiltration Capacity (VIC) process-based hydrologic model to a neural network ($g$) that estimates physical parameters of VIC ($\boldsymbol{\theta}$) using some widely available attributes ($\boldsymbol{A}$): $\boldsymbol{\theta} = g(\boldsymbol{A})$. In an "end-to-end" workflow, $\boldsymbol{\theta}$ is then sent to VIC, whose outputs are compared with observations, effectively turning the parameter calibration problem into a machine learning problem, trained on all sites simultaneously using backpropagation and gradient descent (Figure S1a). As a result of this global loss function, dPL exhibits advantages over traditional calibration on multiple fronts, for three different datasets (soil moisture, CAMELS streamflow, and global headwater runoff). The parameter sets are spatially coherent (Figure S1b-c) and extrapolate better in space (Figure S1d-e). dPL is hyper efficient: a job that normally takes a 100-CPU cluster 2-3 days now takes a single Graphical Processing Unit (GPU) one hour. dPL allows the combined model to output unobserved variables while addressing the notorious problem of parameter equifinality[119]. As summarized earlier, these are the great advantages we expect to harness with differentiable modeling.

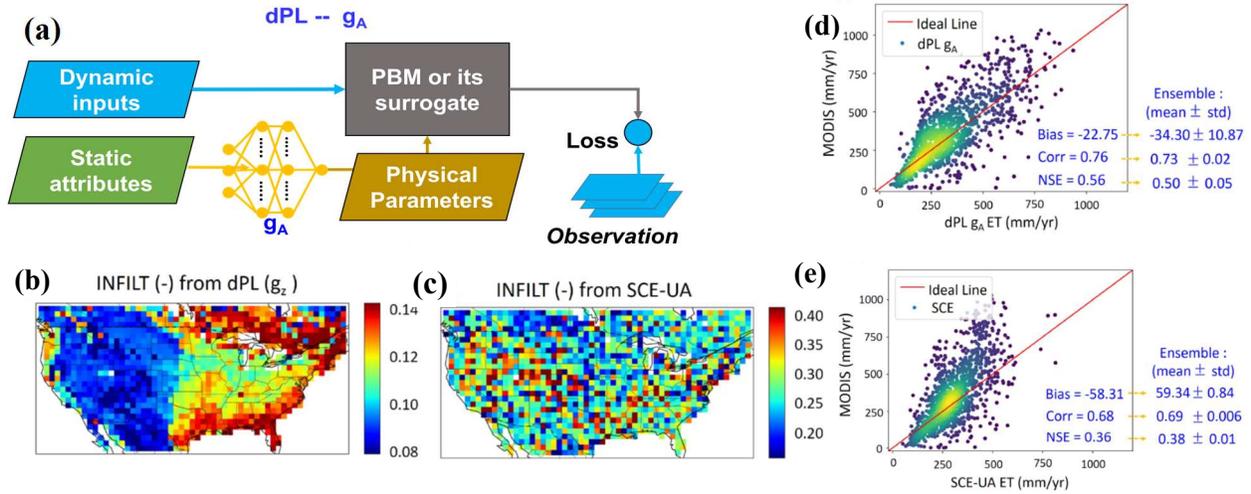

*Figure S1. (From Tsai et al.[100], reprint allowed via Creative Commons Attribution 4.0 International License, http://creativecommons.org/licenses/by/4.0/) (a) Structural diagram of one of the dPL frameworks called $g_A$; (b & c) The estimated infiltrating curve parameter (INFILT) from dPL vs. the site-by-site calibrated shuffled complex evolutionary algorithm (SCE-UA);(d & e) dPL better matches the MODIS satellite product for uncalibrated variable ET than does SCE-UA.*

Example 2. Physics-informed neural networks (PINNs)[101,140], while first published in 2017 before the existence of the term "differentiable geosciences", could be perceived as a genre of DG as the gradient information is critically employed. PINNs pose problems in a unique way, seeking to train a neural network with space-time coordinates as inputs, $h(t,x)$ where $x$ represents spatial coordinates and $t$ is time such that (i) $h(t,x)$ agrees with known data points at $(t,x)$, and (ii) the derivatives $dh/dx$, $dh/dt$, etc. agree with the governing partial differential equations. Physical parameters could also be part of the inputs to the $h$ network[153]. PINNs are a highly innovative approach tested on a large variety of applications in many domains, and there have been a number of good reviews of this work[130,154]. PINNs have made enormous strides, with novel inversion uses such data assimilation[140] and learning governing equations, but, as with other methods, there are also some limitations. Obviously, the function $h(t,x)$ is tied to the initial and boundary conditions so it needs to be trained separately for different initial/boundary condition pairs, and the form of the inputs limits the neural network to certain types (multilayer perceptron network) that are not the easiest to train. However, the learned parameters and constitutive relationships can describe the system under a wide range of boundary and initial conditions. Furthermore, the fidelity of the trained network to physical equations must be carefully examined.

In geosciences, a PINN method for learning unknown parameter fields and constitutive relationships was proposed[104] (Figure S2). As an example, steady-state groundwater flow in an aquifer with an unknown conductivity field and unsaturated flow in the vadose zone with an unknown pressure-dependent conductivity were considered. In the unsaturated flow application, it was assumed that only sparse measurements of pressure head were available. The quantities of interest were the unsaturated conductivity as a function of the pressure head, and the pressure head field. Notably, it was assumed that no measurements of the unknown parameters were available. In the proposed PINN method, both quantities of interest were represented with neural networks (NNs) (with unknown parameters). This step created a differentiable model of the unsaturated flow in the vadose zone. It was also assumed that the pressure head measurements could be described by the steady-state Richards equation. Substituting the NN approximations into this equation formed the axillary residual NN, which shared the (unknown) parameters with the primary NNs. For the primary NNs to satisfy the governing equation, the residual NN should be zero everywhere in the domain – in other words, the exact measurements of the residuals are available everywhere in the domain. The NNs were trained jointly using the pressure head measurements. Since the conductivity and residual NNs share the same parameters, estimating parameters in the residual NN also provides the parameterization of the conductivity NN. Figure S2a shows the reference pressure head field and the locations of the measurements. Figure S2b shows the point errors in the estimated pressure head field. The reference and estimated unsaturated conductivity functions are shown in Figure S2c. These figures demonstrate that the PINN method can learn both the state variable and the constitutive relationship very accurately.

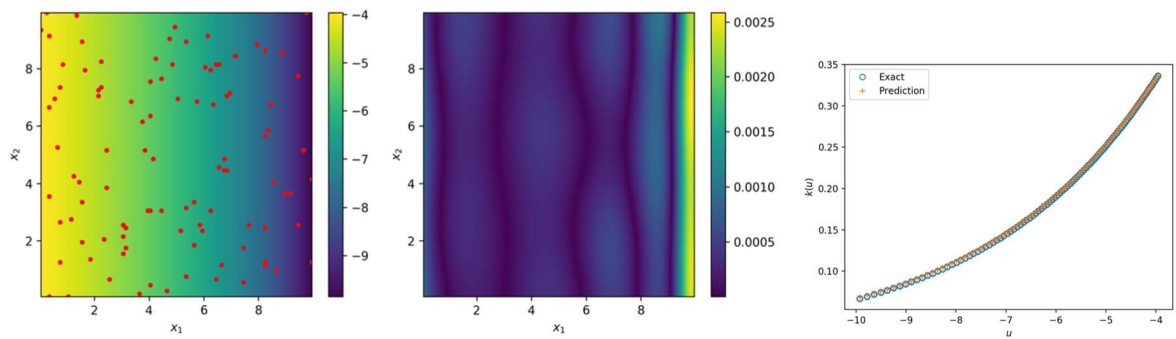

*Figure S2. (from Tartakovsky et al.[104], reprint permission obtained) (a) The reference pressure head field and the locations of the measurements. (b) The point errors in the estimated head field. (c) The reference and estimated conductivities as functions of the pressure head.*

*Disclaimer*

Any use of trade, firm, or product names is for descriptive purposes only and does not imply endorsement by the U.S. Government.

*Supplemental References*

…